\documentclass[acmtog, authorversio]{acmart}
\usepackage{multirow}
\usepackage{graphicx}
\graphicspath{{Figures/}}
\usepackage{subfigure}

\AtBeginDocument{%
  \providecommand\BibTeX{{%
    \normalfont B\kern-0.5em{\scshape i\kern-0.25em b}\kern-0.8em\TeX}}}

\setcopyright{acmcopyright}
\copyrightyear{2018}
\acmYear{2018}
\acmDOI{XXXXXXX.XXXXXXX}

\acmConference[Conference acronym 'XX]{Make sure to enter the correct
  conference title from your rights confirmation emai}{June 03--05,
  2018}{Woodstock, NY}
\acmBooktitle{Woodstock '18: ACM Symposium on Neural Gaze Detection,
 June 03--05, 2018, Woodstock, NY} 
\acmPrice{15.00}
\acmISBN{978-1-4503-XXXX-X/18/06}

\begin{document}

\title{REACT2023: the first Multi-modal Multiple Appropriate Facial Reaction Generation Challenge}


\author{Siyang Song}
\authornote{Both authors contributed equally to this research.}
\authornote{Corresponding author.}
\affiliation{%
  \institution{University of Leicester \& University of Cambridge}
 \city{Leicester}
 \country{United Kingdom}
  }
  \email{ss1535@leicester.ac.uk}
  
\author{Micol Spitale}
\authornotemark[1]
\orcid{0000-0002-3418-1933}
\affiliation{%
  \institution{University of Cambridge}
  \city{Cambridge}
  \country{United Kingdom}
  }
  \email{ms2871@cam.ac.uk}

  \author{Cheng Luo}
\affiliation{%
  \institution{Shenzhen University}
  \city{Shenzhen}
  \country{China}}
\email{Chengluo@cn}

  \author{Germán Barquero}
\affiliation{%
  \institution{Universitat de Barcelona \& Computer Vision Center}
  \city{Barcelona}
  \country{Spain}}
\email{germanbarquero@ub.edu}

  \author{Cristina Palmero}
\affiliation{%
  \institution{Universitat de Barcelona \& Computer Vision Center}
  \city{Barcelona}
  \country{Spain}}
\email{crpalmec7@alumnes.ub.edu}

  \author{Sergio Escalera}
\affiliation{%
  \institution{Universitat de Barcelona \& Computer Vision Center}
  \city{Barcelona}
  \country{Spain}}
\email{sergio@maia.ub.es}

  \author{Michel Valstar}
\affiliation{%
  \institution{University of Nottingham}
  \city{Nottingham}
  \country{United Kingdom}}
\email{michel@blueskeye.com}

  \author{Tobias Baur}
\affiliation{%
  \institution{University of Augsburg}
  \city{Augsburg}
  \country{Germany}}
\email{tobias.baur@uni-a.de}

  \author{Fabien Ringeval}
\affiliation{%
  \institution{Université Grenoble Alpes}
  \city{Grenoble}
  \country{France}}
\email{fabien.ringeval@imag.fr}

\author{Elisabeth André}
\affiliation{%
  \institution{University of Augsburg}
  \city{Augsburg}
  \country{Germany}}
\email{andre@uni-a.de}

\author{Hatice Gunes}
\affiliation{%
  \institution{University of Cambridge}
  \city{Cambridge}
  \country{United Kingdom}}
\email{hatice.gunes@cl.cam.ac.uk}

\renewcommand{\shortauthors}{Trovato and Tobin, et al.}

\begin{abstract}

The Multi-modal Multiple Appropriate Facial Reaction Generation Challenge (REACT2023) is the first competition event focused on evaluating multimedia processing and machine learning techniques for generating human-appropriate facial reactions in various dyadic interaction scenarios, with all participants competing strictly under the same conditions. The goal of the challenge is to provide the first benchmark test set for multi-modal information processing and to foster collaboration among the audio, visual, and audio-visual affective computing communities, to compare the relative merits of the approaches to automatic appropriate facial reaction generation under different spontaneous dyadic interaction conditions. This paper presents: (i) novelties, contributions and guidelines of the REACT2023 challenge; (ii) the dataset utilized in the challenge; and (iii) the performance of baseline systems on the two proposed sub-challenges: Offline Multiple Appropriate Facial Reaction Generation and Online Multiple Appropriate Facial Reaction Generation, respectively. The challenge baseline code is publicly available at \url{https://github.com/reactmultimodalchallenge/baseline_react2023}.

\end{abstract}





\maketitle


\section{Introduction}

The Multi-modal Multiple Appropriate Facial Reaction Generation Challenge (REACT2023) is the first competition aimed at the comparison of multimedia processing and machine learning methods for automatic human appropriate facial reaction generation under different dyadic interaction scenarios, with all participants competing strictly under the same conditions.

As discussed in \cite{song2023multiple}, the generation of human facial reactions in dyadic interactions poses uncertainties, as various (non-verbal) reactions may be deemed appropriate in response to specific speaker behaviours. Although some prior studies~\cite{song2023multiple,huang2017dyadgan,huang2018generating,song2022learning,shao2021personality,barquero2022didn, barquero22comparison, palmero22chalearn} have already explored the task of automatically generating human-style facial and bodily reactions based on the conversational partner's behaviours, they mainly focused on reproducing a specific real facial reactions corresponding to the input speaker behaviour, which introduces challenges due to the potential divergence of non-verbal reaction labels for similar speaker behaviours at the training stage. Just a very recent work \cite{ng2022learning} presents a non-deterministic approach to generate multiple listener reactions from a speaker behaviour but without evaluating the appropriateness of the generated reactions. 
For an in-depth discussion on the multiple different appropriate facial reaction generation task, please refer to Song et al. \cite{song2023multiple}. In other words, none of existing approaches can automatically generate \emph{multiple appropriate reactions} in a dyadic interaction setting, which is a more realistic task, while lacking \textit{objective} measures to evaluate the appropriateness of the generated facial reactions.

The main goal of the REACT2023 challenge is to facilitate collaboration among multiple communities from different disciplines, in particular, the affective computing and multimedia communities and researchers in the psychological and social sciences specializing in expressive facial behaviours. The challenge aims to encourage \textbf{the initial development and benchmarking} of Machine Learning (ML) models capable of generating \textit{appropriate} facial reactions in response to a given stimulus, using three state-of-the-art datasets for dyadic interaction research, namely, RECOLA \cite{ringeval2013introducing}, NOXI \cite{cafaro2017noxi}, and UDIVA \cite{palmero2021context}. As a part of the challenge, we will provide challenge participants with the REACT2023 Challenge Dataset, comprising segmented 30-second interaction video clips (video pairs) from the aforementioned three datasets, annotated with challenge-specific labels indicating the appropriateness of facial reactions. 
We will then invite the participating groups to submit their developed / trained ML models for evaluation, which will be benchmarked in terms of the appropriateness, diversity, and synchrony of the generated facial reactions.

The main contributions and novelties are introduced for the REACT2023 with two separated sub-challenges focusing on online and offline appropriate facial reaction generations as: 

\begin{itemize}
    \item \textbf{Offline Multiple Appropriate Facial Reaction Generation (Offline MAFRG)} task focuses on generating multiple appropriate facial reaction videos from the input speaker behaviour (i.e., audio-visual clip). 
    Specifically, this task aims to develop a machine learning model $\mathcal{H}$ that takes the entire speaker behaviour sequence $B_S^{t_1,t_2}$ as the input, and generates multiple ($M$) appropriate and realistic / naturalistic spatio-temporal facial reactions $p_f(b_S^{t_1,t_2})_1, \cdots, p_f(b_S^{t_1,t_2})_M$; where $p_f(b_S^{t_1,t_2})_m$ is a multi-channel time-series -- consisting of AUs, facial expressions, valence and arousal state -- which represent the $m_{th}$ predicted appropriate facial reaction in response to $B_S^{t_1,t_2}$. 
    Based on the predicted facial attributes, the challenge participants have to generate $M$ appropriate and realistic / naturalistic spatio-temporal facial reactions (2D face image sequences) given each input speaker behaviour.

    \item \textbf{Online Multiple Appropriate Facial Reaction Generation (Online MAFRG)} task focuses on the continuous generation of facial reaction frames based on current and previous speaker behaviours. 
    This task aims to develop a machine learning model $\mathcal{H}$ that estimates multiple facial attributes (AUs, facial expressions, valence and arousal state) representing each appropriate facial reaction frame (i.e.,  $\gamma_\text{th} \in [t_1,t_2]$ frame) by only considering the $\gamma_\text{th}$ frame and its previous frames of the corresponding speaker behaviour (i.e., ${t_1}_\text{th}$ to $\gamma_\text{th}$ frames in $B_S^{t_1,t_2}$), rather than taking all frames from ${t_1}$ to ${t_2}$ into account. 
    The model is expected to gradually generate multiple multi-channel facial attribute time-series to represent all face frames of multiple appropriate and realistic / naturalistic spatio-temporal facial reactions $p_f(b_S^{t_1,t_2})_1, \cdots, p_f(b_S^{t_1,t_2})_M$, where $p_f(b_S^{t_1,t_2})_m$, where $p_f(b_S^{t_1,t_2})_m$ is a multi-channel time-series -- consisting of AUs, facial expressions, valence and arousal state -- representing the $m_{th}$ predicted appropriate facial reaction in response to $B_S^{t_1,t_2}$. 
    Based on the predicted facial attributes, the challenge participants have to generate $M$ appropriate and realistic / naturalistic spatio-temporal facial reactions (2D face image sequences) given each input speaker behaviour.
\end{itemize}

While participants are welcome to report their results obtained on the validation partition, they are restricted to a maximum of five submission attempts per sub-challenge for presenting their results on the test partition. Both sub-challenges allow participants to explore their own features and machine learning algorithms. We additionally provide standardized audio-visual feature sets (Sec.~\ref{sec: baseline-feature}), along with the baseline system scripts available in a public repository\footnote{\url{https://github.com/reactmultimodalchallenge/baseline_react2023/tree/main}}, to facilitate the reproduction of baseline features and facial reaction generation systems (Sec. \ref{sec: baseline-system}). All participants are required to report their results achieved on the validation and test partitions.

The REACT2023 Challenge adopts the metrics defined in \cite{song2023multiple} to evaluate the performance of the submitted models in terms of generated facial reactions, namely: appropriateness, diversity, realism and synchrony. Participants are required to submit their developed model and checkpoints, which are evaluated and visualised based on the framework provided by \cite{song2023multiple}. 
The ranking of the submitted model competing in the Challenge relies on the two metrics: Appropriate facial reaction distance (FRDist) and facial reactions' diverseness FRDiv, for both sub-challenges. 
In addition, Facial reaction correlation (FRCorr), Facial reaction realism (FRRea), Facial reaction variance (FRVar), Diversity among facial reactions generated from different speaker behaviours (FRDvs) and Synchrony between generated facial reactions and speaker behaviours (FRSyn) should also be reported.

To be eligible to participate in the Challenge, each team must fulfill specific criteria, including the submission of thoroughly explained source code, well-trained models and associated checkpoints, accompanied by a paper submitted to the REACT2023 Challenge describing the proposed methodology and the achieved results. 
These papers undergo a rigorous peer-review by the REACT2023 challenge technical program committee. Only contributions that meet the terms and conditions\footnote{https://sites.google.com/cam.ac.uk/react2023/home} requirements are eligible for participation. The organisers do not engage in active participation themselves, but instead undertake a re-evaluation of the findings from the best performing systems in each sub-challenge.

The remainder of this paper is organised as follows. The relevant related works are reviewed in Sec. \ref{sec:related}. We then introduce the Challenge corpora in Sec. \ref{sec:Challenge_corpora}, and the evaluation metrics in Sec. \ref{sec:metrics}. The baseline audio-visual feature sets, and baseline facial reaction generation systems are introduced in Sec. \ref{sec: baseline-system}, respectively. We finally conclude the challenge in Sec. \ref{sec: conclusion}.

\section{Related Work}
\label{sec:related}

In this section, we first review previous works on automatic facial reaction generation in Sec. \ref{subsec:facial reaction generation}, and then further summarises common facial reaction visualization strategies in Sec. \ref{subsec:facial reaction visualization}.

\subsection{Facial reaction generation}
\label{subsec:facial reaction generation}

As discussed in \cite{song2023multiple}, in dyadic interactions, human listeners could express a broad spectrum of appropriate non-verbal reactions for responding to a specific speaker behaviour. However, most prior works have attempted to reproduce the listener's real facial reaction that corresponds to the input speaker behaviour from a deterministic perspective~\cite{barquero2022didn}. For example, Huang et al. \cite{huang2017dyadgan} trained a conditional Generative Adversarial Network (GAN) \cite{mirza2014conditional,goodfellow2020generative} and attempted to generate the listener's real facial reaction sketch from the corresponding speaker's facial action units (AUs). Similar frameworks \cite{huang2018generating,huang2018generative,nojavanasghari2018interactive, woo2021creating, woo2023amii} have been extended for the same purpose (i.e., reproducing the specific real facial reaction from each input speaker behaviour), where more modalities (e.g., low-level facial expression features and audio features) are employed as the input. In particular, Song et al. \cite{song2022learning,shao2021personality} propose to explore a person-specific network for each listener, and thus each listener's person-specific facial reactions could be reproduced. Other works have explored the generation of other non-verbal behaviours, such as hand gestures, posture, and facial reaction altogether in face-to-face scenarios~\cite{palmero22chalearn,barquero22comparison,tuyen22context}. They all highlighted that avoiding the convergence to a mean reaction is challenging with existing deterministic approaches. However, the training process of such deterministic approaches would face the ill-posed 'one-to-many mapping' problem (i.e., one speaker behaviour corresponds to one appropriate facial reaction distribution, and even the same listener can express different facial reactions in response to the same speaker behaviour under different contexts), making them theoretically impossible to learn good hypothesis \cite{song2023multiple}.

Recently, a few works have started to explore the non-deterministic perspective of this problem, which can predict different facial reactions from the same input. For example, Jone et al. \cite{jonell2020let} proposed an architecture that is able to sample multiple avatar's facial reaction to the interlocutor's speech and facial motion. Similarly, \cite{ng2022learning} presented a VQ-VAE-based multimodal method that leverages the speaker's behaviour (i.e., speech features and facial motion). This model can also generate multiple listener's facial reactions from the same input speaker behaviour, despite it does not consider the appropriateness of the generate facial reactions. Geng et al. \cite{geng2023affective} proposed to exploit pre-trained large language models and vision-language models together to retrieve the best listener's reaction to the speaker's speech. Their method allows the user to control the reaction retrieval process with textual instructions. However, reactions can only be retrieved from a pre-existing video database, and therefore limited to the identities and reactions available in the database. Xu et al. \cite{xu2023reversible} and Luo et al. \cite{luo2023reactface} recently proposed a reversible graph neural network-based and a transformer-based models, respectively. Both of them reformulated the 'one-to-many mapping' problems (i.e., one input speaker behaviour could corresponds to multiple appropriate facial reaction labels) occurring in the facial reaction generation models' training into 'one-to-one mapping' problem. Consequently, at the inference stage, multiple different but appropriate facial reactions could be sampled from the learned distribution.


\subsection{Facial reaction visualization}
\label{subsec:facial reaction visualization}


A common strategy for visualising facial behaviours is through using 3D morphable models. For example, Ng et al. \cite{ng2022learning} proposed to use 3D Morphable Model (3DMM) coefficients to represent and visualize facial reactions, which were then transformed to a 2D image with a proprietary software. Xing et al. \cite{xing2023codetalker} proposed to discretize the continuous space of 3DMM coefficients using a codebook learnt by using a VQ-VAE. Thanks to the mapping to a finite discrete space, the uncertainty of the facial generation is significantly reduced, yielding higher quality results. The similar 3DMM coefficient-based strategy is also used by Zhou et al. \cite{zhou2022responsive}, whose approach only aimed to reproduce the real facial reaction in terms of three emotional states (positive, neutral, and negative) rather than detailed facial muscle movements.

Meanwhile, 2D facial behaviours are frequently visualised based on Generative Adversarial Networks (GANs) ~\cite{chen2019realistic} conditioned on the predicted facial expression latent representations, where facial image sequences are generated based on manually defined conditions such pre-defined AUs \cite{ganimation}, 2D facial landmarks \cite{otberdout2020dynamic} and audio signals \cite{fan2022joint,richard2021meshtalk} without considering interaction scenarios (i.e., they do not predict reactions from speaker behaviours). Moreover, recent studies are also proposed to generate 2D facial image sequence from 3D facial behaviours. For example, the PIRender \cite{ren2021pirenderer} and FaceVerseV2 \cite{wang2022faceverse} frameworks can translate 3DMM coefficient to 2D facial images conditioned on the portrait of reference identity, where the FaceVerseV2 framework is also employed in this paper to generate facial reaction image sequences.

\section{Challenge Corpora}
\label{sec:Challenge_corpora}

The REACT2023 Challenge relies on three corporas: NoXi \cite{cafaro2017noxi}, UDIVA \cite{palmero2021context}, and RECOLA \cite{ringeval2013introducing} datasets. We provide a short overview of each dataset below and recommend readers to check the original work for details.

\subsection{Employed datasets}

\subsubsection{NOvice eXpert Interaction dataset}

The NOvice eXpert Interaction (NOXI) is a dyadic interaction dataset that is annotated during an information retrieval task targeting multiple languages, multiple topics, and the occurrence of unexpected situations. NoXi is a corpus of screen-mediated face-to-face interactions recorded at three locations (France, Germany and UK), spoken in seven languages (English, French, German, Spanish, Indonesian, Arabic and Italian) discussing a wide range of topics.

\subsubsection{Understanding Dyadic Interactions from Video and Audio signals dataset} The UDIVA dataset features face-to-face interactions between pairs of participants performing a set of collaborative and competitive tasks, using one of the three languages included (i.e., English, Spanish or Catalan). We rely on the UDIVA v0.5 data subset~\cite{palmero22chalearn}, composed of 145 dyadic interaction sessions between 135 participants, with a total of 80 hours of recordings. Each clip contains two audio-visual files that record the dyadic interaction between a pair of participants, as well as the conversation transcripts, and metadata about the participants, sessions, and tasks (e.g., sociodemographics, internal state, self-reported personality, relationship among participants, task difficulty).

\subsubsection{REmote COLlaborative and Affective dataset}
The REmote COLlaborative and Affective (RECOLA) database consists of 9,5 hours of audio, visual, and physiological (electrocardiogram, and electrodermal activity) recordings of online dyadic interactions between 46 French speaking participants, who were solving a task in collaboration.

\subsection{Appropriate Facial Reaction (AFR) dataset}


We first segmented the audio-video data of all the three datasets in 30-seconds long clips as in \cite{ambady1992thin}. Then, we cleaned the dataset by selecting only the dyadic interaction with complete data of both conversational partners (where both faces were in the frame of the camera). This resulted into 8616 clips of 30 seconds each (71,8 hours of audio-video clips), specifically: 5870 clips (49 hours) from the NoXi dataset
, 54 clips (0,4 hour) from the RECOLA dataset
, and 2692 clips (22,4 hours) from the UDIVA dataset.

We divided the datasets into training, test and validation sets. Specifically, we split the datasets with a subject-independent strategy (i.e., the same subject was never included in the train and test sets). 
We also attempted to balance the language across the training, test and validation sets. However, since many users interacted in multiple sessions, we were not able to get a language-balance split. 
This results in a training composed by: 1030 video clips -- 8,6 hours -- of UDIVA, 1585 video clips -- 13,2 hours -- of NOXI, and 9 video clips -- 0,1 hour -- of RECOLA. The test set was composed by: 39 video clips -- 0,3 hour -- of UDIVA, 797 video clips -- 6,6 hours -- of NoXI, and 9 video clips -- 0,1 hour -- of RECOLA. While the validation set has the following interactions: 277 video clips -- 2,3 hours -- of UDIVA, 553 video clips -- 4,6 hours -- of NoXI, and 9 video clips -- 0,1 hour -- of RECOLA. Tables \ref{tab:train-set}, \ref{tab:test-set}, and \ref{tab:val-set} collect the details about the training, testing, and validation sets.

\begin{table}[]
\footnotesize
\caption{Data description of the training set from the UDIVA, NoXi, and RECOLA datasets.}
\label{tab:train-set}
\resizebox{\columnwidth}{!}{%
\begin{tabular}{lccc}
\hline
 &
  \multicolumn{1}{c}{\textbf{NoXI}} &
  \multicolumn{1}{c}{\textbf{UDIVA}} &
  \multicolumn{1}{c}{\textbf{RECOLA}}\\
  &\multicolumn{1}{c}{\textbf{(\# clips, hours)}} &
  \multicolumn{1}{c}{\textbf{(\# clips, hours)}} &
  \multicolumn{1}{c}{\textbf{(\# clips, hours)}}\\
  \hline
Clips     & 1585, 13.2 h & 1030, 8.6 h & 9,  0.1\\
\hline
English    & 972, 8.1 h  & 66, 0.6 h   & - \\
Catalan   & -   & 211, 1.8 h  & - \\
Spanish    & 144, 1.2 h  & 753, 6.3 h  & - \\
Arabic    & -   & -    &- \\
Italian   & 42, 0.4 h   & -    &- \\
Indonesian & 151, 1.3 h  & -    & - \\
German     & -   &-   &- \\
French     & 276, 2.3 h  & -   & 9, 0.1 h \\
\hline
\end{tabular}%
}
\end{table}

\begin{table}[]
\caption{Data description of the testing set from the UDIVA, NoXi, and RECOLA datasets.}
\label{tab:test-set}
\resizebox{\columnwidth}{!}{%
\begin{tabular}{lccc}
\hline
 &
  \multicolumn{1}{c}{\textbf{NoXI}} &
  \multicolumn{1}{c}{\textbf{UDIVA}} &
  \multicolumn{1}{c}{\textbf{RECOLA}}\\
  &\multicolumn{1}{c}{\textbf{(\# clips, hours)}} &
  \multicolumn{1}{c}{\textbf{(\# clips, hours)}} &
  \multicolumn{1}{c}{\textbf{(\# clips, hours)}}\\
  \hline
Clips   & 797, 6.6 h & 39, 0.3 h & 9, 0.1 h \\
\hline
English   & -   & -  & - \\
Catalan    & -   & -  & - \\
Spanish    & -   & 39, 0.3 h & - \\
Arabic     & -   & -  & - \\
Italian    & -   & -  & - \\
Indonesian & -   & -  & - \\
German     & 486, 4.1 h & -  & - \\
French     & 311, 2.6 h & - & 9, 0.1 h \\
\hline
\end{tabular}%
}
\end{table}

\begin{table}[]
\caption{Data description of the validation set from the UDIVA, NoXi, and RECOLA datasets.}
\label{tab:val-set}
\resizebox{\columnwidth}{!}{%
\begin{tabular}{lrrrrrr}
\hline
 &
   \multicolumn{1}{c}{\textbf{NoXI}} &
  \multicolumn{1}{c}{\textbf{UDIVA}} &
  \multicolumn{1}{c}{\textbf{RECOLA}}\\
  &\multicolumn{1}{c}{\textbf{(\# clips, hours)}} &
  \multicolumn{1}{c}{\textbf{(\# clips, hours)}} &
  \multicolumn{1}{c}{\textbf{(\# clips, hours)}}\\
  \hline
Clips     & 553, 4.6 h & 277, 2.3 h & 9.0, 0.1 h \\
\hline
English    & -   & 121, 1 h & - \\
Catalan    & -   & -   & - \\
Spanish   & -   & 156, 1.3 h & -\\
Arabic     & 47, 0.4 h  & -   & - \\
Italian    & 31, 0.3 h  & -   & - \\
Indonesian & -   & -   & - \\
German     & 181, 1.5 h & -   & - \\
French    & 294, 2.5 & -  & 9.0, 0.1 h \\
\hline
\end{tabular}%
}
\end{table}

\section{Evaluation Metrics}
\label{sec:metrics}

In this challenge, we ask participants to develop models that can generate two types of outputs representing each generated facial reaction: (i) 25 facial attribute time-series (explained in Sec. \ref{sec: baseline-feature}); and (ii) 2D and 3D facial image sequence. 

We follow \cite{song2023multiple} to comprehensively evaluate four aspects of the facial reactions generated by participant models. In particular, three aspects are assessed based on the 25 facial attribute time-series: (i) \textbf{Appropriateness}, which is evaluated using two metrics, namely Dynamic Time Warpping (DTW) and Concordance Correlation Coefficient (CCC). Both metrics are computed between the generated facial reactions and its most correlated appropriate real facial reaction, which are referred as \textbf{FRDist} and \textbf{FRCorr}, respectively; (ii) \textbf{Diversities}, which encompass inter-condition and inter-frame variations. Metrics such as \textbf{FRVar}, \textbf{FRDiv}, and \textbf{FRDvs}, as defined in \cite{song2023multiple}, are employed to measure these diversities; and (iii) \textbf{Synchrony}, which examines the alignment between the generated facial reactions and the corresponding speaker behaviour. The Time Lagged Cross Correlation (TLCC) is employed as a metric for measuring this synchrony, referred to as \textbf{FRSyn} in this challenge.
Based on the generated 2D facial image sequence, we also evaluate the (iv) \textbf{Realism} of the generated facial reactions, which is assessed using the Fréchet Inception Distance (FID) between the distribution of the generated facial reactions and the distribution of the corresponding appropriate real facial reactions, denoted as \textbf{FRRea}.

\section{Baselines}
\label{sec: baseline-system}

This section presents the baseline systems developed for the REACT23 Challenge.
First, we detail the audio and visual behavioural features extracted which are used for describing facial reactions in the evaluation protocol (Sec. \ref{sec: baseline-feature}). Then, we propose two baseline systems in Sec. \ref{subsec:baseline systems}. Finally, we report all results achieved by our baseline systems in Sec. \ref{subsec:baseline results}.

\subsection{Behavioural features}
\label{sec: baseline-feature}

\subsubsection{Visual features}

We follow \cite{song2023multiple} to provide three widely-used frame-level facial attribute features for each video frame as the baseline facial features. This includes the occurrence of $15$ facial action units (AUs), $2$ facial affect -- valence and arousal intensities -- and the probabilities of $8$ categorical facial expressions. Specifically, 15 AUs' occurrence (AU1, AU2, AU4, AU6, AU7, AU9, AU10, AU12, AU14, AU15, AU17, AU23, AU24, AU25 and AU26) are predicted by the state-of-the-art GraphAU model \cite{luo2022learning,song2022gratis}, while the facial affects (i.e., valence and arousal intensities) and 8 facial expression probabilities (i.e., Neutral, Happy, Sad, Surprise, Fear, Disgust, Anger and Contempt) are predicted using the approach proposed by \cite{toisoul2021estimation}.

\subsubsection{Audio features}

We also apply OpenSmile \cite{eyben2010opensmile} to extract clip-level audio descriptors, including GEMAP and MFCC features. Consequently, we represent each speaker behaviour by combining all frame-level descriptors as a multi-channel audio-visual time-series behavioural signal.



\subsection{Baseline systems}
\label{subsec:baseline systems}


\begin{figure}
\centering
\subfigure[Illustration of the offline Trans-VAE baseline]{
\includegraphics[width=0.96\columnwidth]{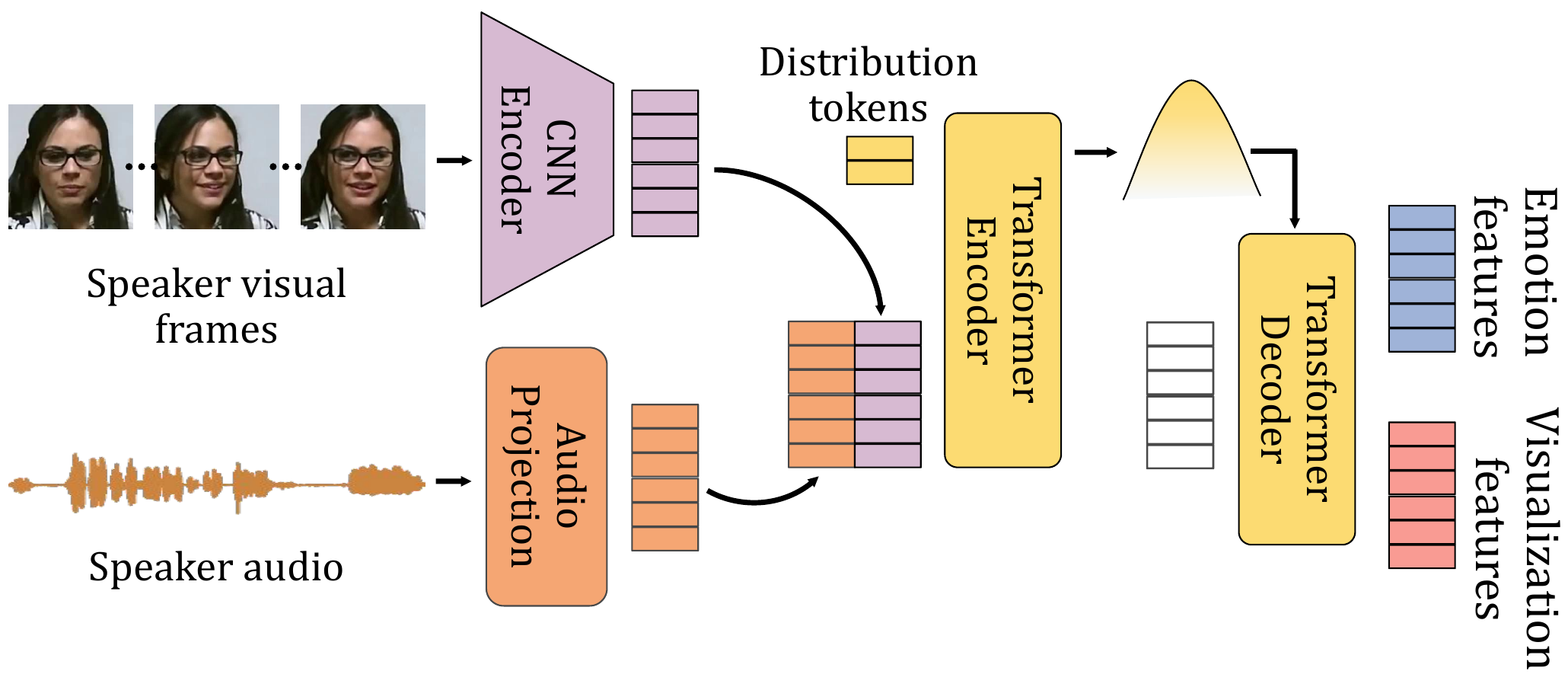} 
}
\subfigure[Illustration of the online Trans-VAE baseline]{
\includegraphics[width=0.96\columnwidth]{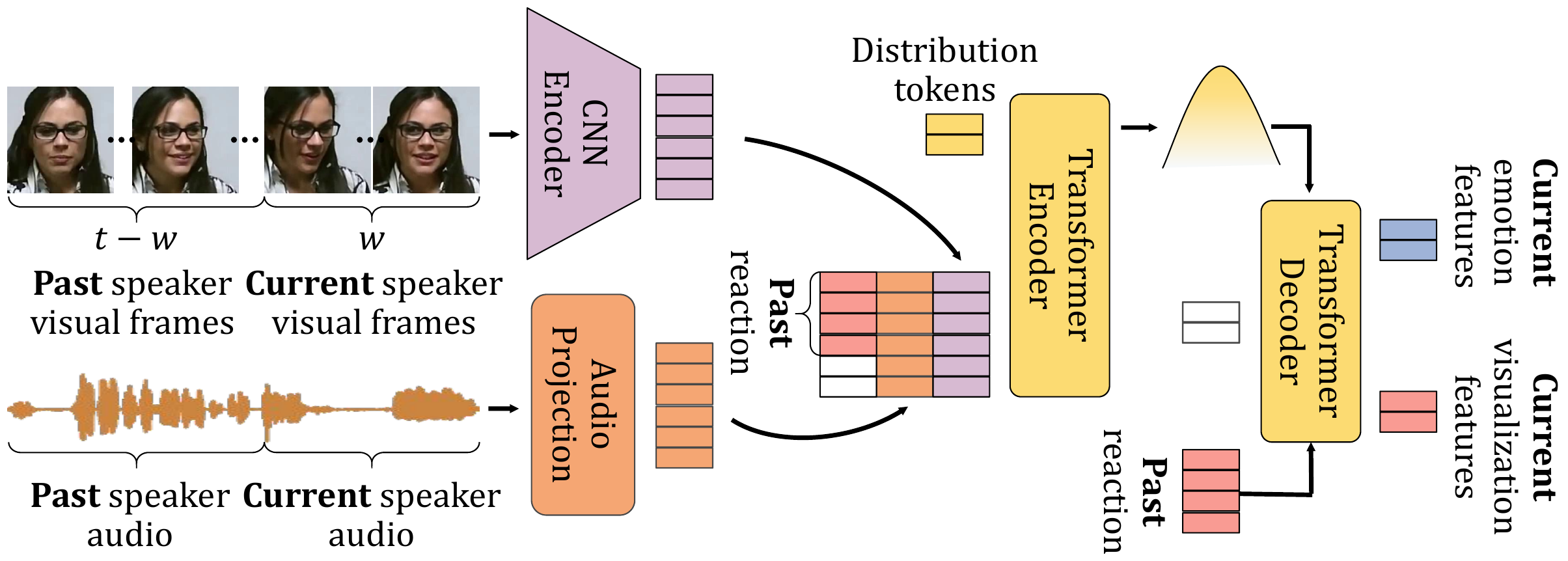} 
}
\DeclareGraphicsExtensions.
\vspace{-0.15in}
\caption{Overview of the Trans-VAE baseline.}
\label{fig:overview_TransVAE}
\end{figure}

\begin{figure}
    \centering
    \includegraphics[width=0.96\columnwidth]{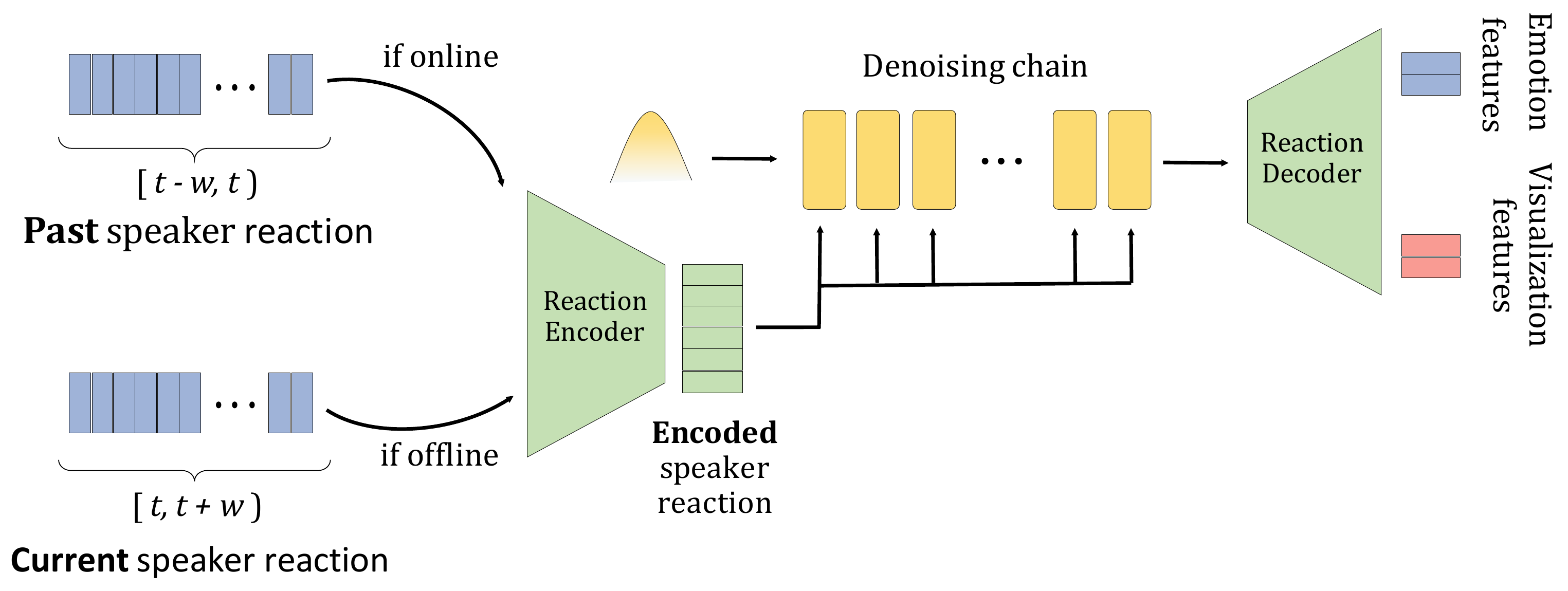}
    \caption{Overview of offline and online BeLFusion baselines. The reaction encoder and decoder are previously trained as a variational autoencoder to learn a lower-dimensional representation of reactions sequences of length $w$. Then, a latent diffusion model is trained to conditionally sample reactions from it. The condition is either the past or the current speaker reaction, for the online and offline approaches, respectively.}
    \label{fig:overview_belfusion}
\end{figure}

\begin{figure*}[htb!]
    \centering
    \includegraphics[width=1.5\columnwidth]{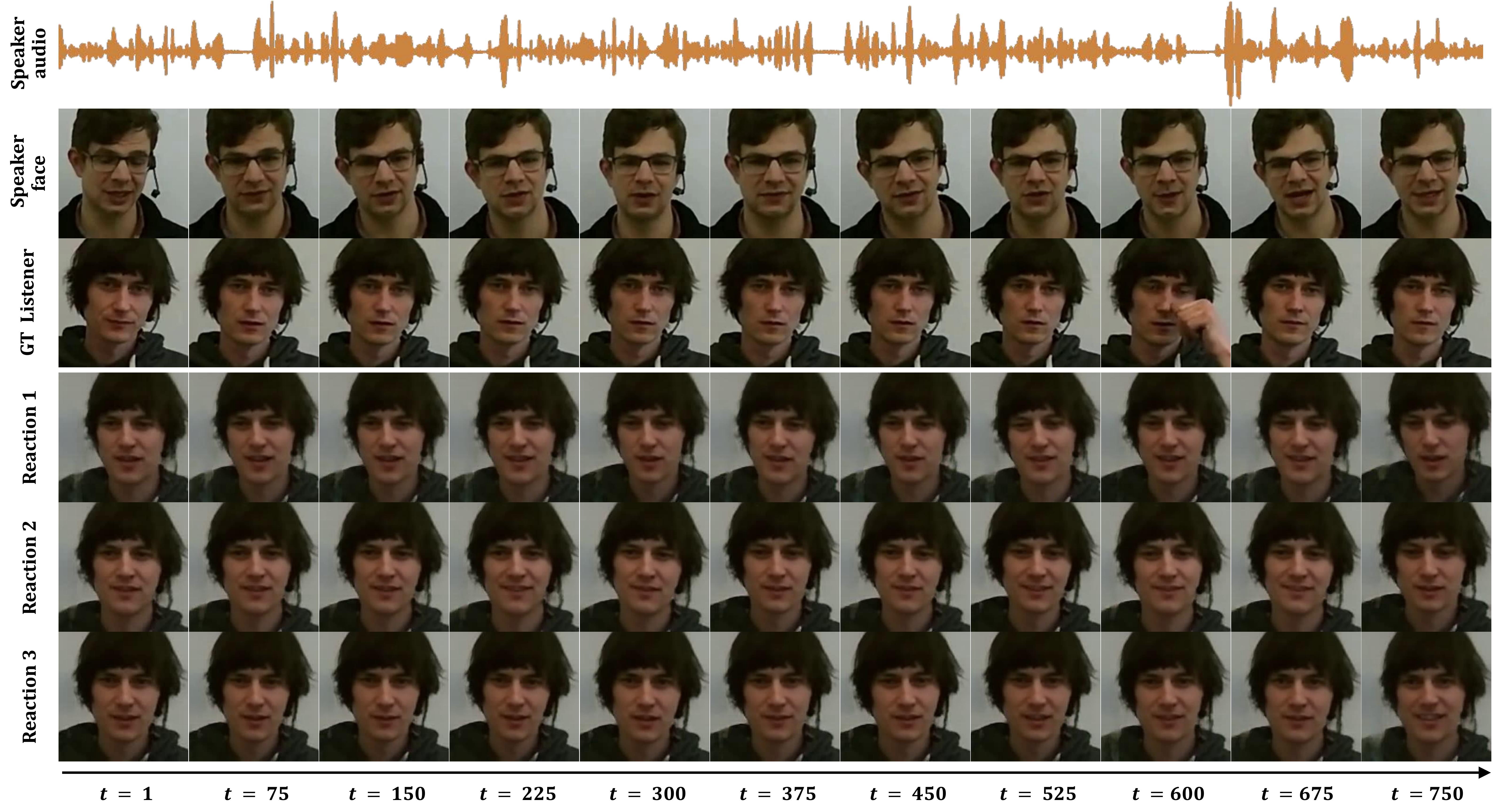} 
        \caption{Examples of generated multiple listener reactions to a given speaker behaviour (including the speaker's audio and face frames). These reactions are generated by an offline Trans-VAE model.  }
    \label{fig:ex_vis_offline}
\end{figure*}

\begin{table*}[t]\centering
 \footnotesize
 \caption{\label{tb:val_quantitative_offline} 
 Baseline offline facial reaction generation results achieved on the validation set.} 
\begin{tabular}{lccccccc}
\toprule 
\multirow{2}{*}{\textbf{Method}} & \multicolumn{2}{c}{\textbf{Appropriateness}}  & \multicolumn{3}{c}{\textbf{Diversity}} &\textbf{Realism}  &   \textbf{Synchrony}\\
\cmidrule(r){2-3} \cmidrule(r){4-6}  \cmidrule(r){7-7}  \cmidrule(r){8-8} 
    & \textbf{FRC} ($\uparrow$) &  \textbf{FRD} ($\downarrow$) &  \textbf{FRDiv} ($\uparrow$) &  \textbf{FRVar} ($\uparrow$) &  \textbf{FRDvs} ($\uparrow$)  & \textbf{FRRea} ($\downarrow$)    & \textbf{FRSyn} ($\downarrow$)\\ 
    \midrule
    GT & 8.42 & 0.00 & 0.0000 & 0.0666 & 0.2251 & 44.31 & 48.52 \\
    \midrule
    B\_Random & 0.04 & 229.37 & 0.1667 & 0.0833 & 0.1667 & - & 46.65 \\
    B\_Mime & 0.35 & 91.37 & 0.0000 & 0.0666 & 0.2251 & - & 44.53 \\
    B\_MeanSeq & 0.01 & 102.17 & 0.0000 & 0.0000 & 0.0000 & - & 47.19 \\
    B\_MeanFr & 0.00 & 102.84 & 0.0000 & 0.0000 & 0.0000 & - & 49.00 \\
    \midrule
    Trans-VAE w/o visual modality & 0.09 & 99.50  & 0.0236  & 0.0030 & 0.0287 & 99.62 & 49.00 \\
    Trans-VAE w/o audio modality & 0.10 & 99.31 & 0.0104 & 0.0015 & 0.0129 & 94.06 & 49.00 \\

    \midrule
    Trans-VAE   & 0.12 & 101.37  &  0.0265 &  0.0045 & 0.0322 & 101.58
    &49.00 \\
    BeLFusion ($k$=1) & 0.12 & 94.39 & 0.0087 & 0.0058 & 0.0106 & - & 47.14 \\
    BeLFusion ($k$=10) & 0.14 & 94.26 & 0.0134 & 0.0078 & 0.0149 & - & 46.94  \\
    BeLFusion ($k$=10) + Binarized AUs & 0.12 & 97.17 & 0.0323 & 0.0173 & 0.0341 & - & 49.00 \\ 
\bottomrule
\end{tabular}
\end{table*}

In this challenge, we first establish a set of naive baselines, namely B\_Random, B\_Mime, and B\_MeanSeq/B\_MeanFr. 
Specifically, B\_Random randomly samples $\alpha = 10$ facial reaction sequences from a Gaussian distribution. B\_Mime generates facial reactions by mimicking the corresponding speaker's facial expressions.  For B\_MeanSeq and B\_MeanFr, the generated facial reactions are decided by the sequence- and frame-wise average reaction in the training set, respectively. Despite their simplicity, these baselines illustrate the bounds of the metrics. For the implementation details and open-source code of all baselines, please refer to our GitHub page at \url{https://github.com/reactmultimodalchallenge/baseline_react2023}.

\textbf{Trans-VAE.} The Trans-VAE baseline has a similar architecture as the TEACH proposed in \cite{athanasiou2022teach}, which consists of: (i) a \textbf{CNN encoder} that encodes the speaker facial image sequence (i.e., a short video) as a sequence-level embedding; (ii) a \textbf{transformer encoder} that first combines learned facial embeddings and audio embeddings (78-dimnesional MFCC features) extracted by Torchaudio library \cite{yang2022torchaudio}, and then predicts a pair of tokens $\mu_\text{token}$ and $\sigma_\text{token}$ representing the Gaussian Distribution of multiple appropriate facial reactions of the corresponding input speaker behaviour, based on not only the combined audio-visual embedding but also a pair of learnable tokens; and (iii) a \textbf{transformer decoder} that samples a set of representations describing an appropriate facial reaction based on the predicted distribution tokens, which include a set of 3D Morphable Model (3DMM) coefficients (i.e., 52 facial expression coefficients, 3 pose coefficients and 3 translation coefficients defined by \cite{wang2022faceverse}) and a emotion matrices (i.e., 25-channel time-series including 15 frame-level AUs' occurrence, 8 frame-level facial expression probabilities as well as frame-level valence and arousal intensities). Based on the learned 3DMM coefficients and the corresponding listener's portrait, FaceVerseV2 \cite{wang2022faceverse} is finally employed to translate the learned 3DMM coefficients to the facial reaction image sequence. 

As illustrated in Fig~\ref{fig:overview_TransVAE}, we apply the Trans-VAE model to both offline and online facial reaction generation sub-challenges. For the offline sub-challenge, it takes the entire sequence of speaker audio-visual behaviours (i.e., $750$ frames corresponding to 30s clip in this challenge) as the input and generates a sequence of facial reactions consisting of $750$ frames. The online Trans-VAE baseline follows \cite{luo2023reactface} to iteratively predict a short segment consisting of $w$ facial reaction frames corresponding to the time $[t-w+1:t]$, where causal mask \cite{luo2023reactface, fan2022faceformer,dong2019unified,press2021train} is employed to avoid future speaker behaviours to be used for the facial reaction prediction. In particular, the $\tau_\text{th}$ facial reaction frame is predicted based on: (i) $t-w$ frames ($[1:t-w]$) of past speaker behaviours; (ii) $t-w$ frames ($[1:t-w]$) of previously predicted facial reactions; and (iii) $\tau$ frames ($[t-w+1:\tau]$) of the current speaker behaviour. The Trans-VAE models for both sub-challenges are trained end-to-end with maximum 50 epochs, where Mean Square Error (MSE) loss function is employed for the 2D facial frame reconstruction; a diversity loss \cite{yuan2020dlow,luo2023reactface} is leveraged to increase sampling diversity; and a KL divergence loss is used to constrain the predicted distribution tokens. To optimize the model, AdamW optimizer \cite{loshchilov2017decoupled} with a fixed learning rate of $1e-5$ is used.

\textbf{BeLFusion.} We use BeLFusion as our second baseline \cite{barquero2022belfusion}, see~\autoref{fig:overview_belfusion}. For the sake of simplicity, we use the version without behavioural disentanglement. BeLFusion is trained in two stages. First, a variational autoencoder (VAE) is trained to learn a lower representation of the visual features (e.g., AUs, facial affects, and expressions) of $w$ frames. On the VAE's head, we include a regressor that transforms the decoded reaction to a sequence of 3DMM coefficients. The VAE losses consists of the KL divergence, the reaction MSE, and the 3DMM coefficients MSE, with weights of $1e-5$, 1, and 1, respectively. We chose $w=50$, and the latent space has dimension 128. The model is trained with the AdamW optimizer~\cite{loshchilov2017decoupled} with a fixed learning rate of 0.001 and weight decay of 0.0005, for 1000 epochs.
In the second stage, a latent diffusion model (LDM) learns to, given the speaker's reaction, predict the lower-dimensional representation of the listener's appropriate facial reaction. Similarly to Trans-VAE, this baseline also adopts a window-based approach where $T/w$ reactions are predicted independently. Then, the $w$-frames-long reactions are stacked to build the full reaction. For the online sub-challenge, the generation of the listener's visual features for the window $[t, t+w)$ is conditioned on the past speaker's features at $[t-w, t)$. It predicts all zeroes for segment $[0,w)$. For the offline sub-challenge, such generation is conditioned on the speaker's features on the same time period: $[t, t+w)$. The LDM's loss is the average of the MSE in the latent space and the MSE in the reconstructed space. The denoising chain has 10 steps, and every denoising step is implemented with a sequence of residual MLPs as in \cite{preechakul2022diffusion}. It is also optimized with the AdamW~\cite{loshchilov2017decoupled}, a learning rate of 0.0001, and a weight decay of 0.0005, for 100 epochs. We include two versions of the model: with $k=1$ and $k=10$. As explained in \cite{barquero2022belfusion}, higher values for $k$ lead to a stronger implicit diversity loss, and therefore more diverse reactions generation.

%

%

\vspace{-3mm}
\subsection{Baseline results}
\label{subsec:baseline results}

\begin{table*}[t]\centering
 \footnotesize
 \caption{\label{tb:test_quantitative_offline} 
Baseline offline facial reaction generation results achieved on the test set.} 
\begin{tabular}{lccccccc}
\toprule 
\multirow{2}{*}{\textbf{Method}} & \multicolumn{2}{c}{\textbf{Appropriateness}}  & \multicolumn{3}{c}{\textbf{Diversity}} &\textbf{Realism}  &   \textbf{Synchrony}\\
\cmidrule(r){2-3} \cmidrule(r){4-6}  \cmidrule(r){7-7}  \cmidrule(r){8-8} 
    & \textbf{FRC} ($\uparrow$) &  \textbf{FRD} ($\downarrow$) &  \textbf{FRDiv} ($\uparrow$) &  \textbf{FRVar} ($\uparrow$) &  \textbf{FRDvs} ($\uparrow$)  & \textbf{FRRea} ($\downarrow$)    & \textbf{FRSyn} ($\downarrow$)\\ 
    \midrule
    GT & 8.74 & 0.00 & 0.0000 & 0.0723 & 0.2474 & 47.50 & 47.72 \\
    \midrule
    B\_Random & 0.04 & 237.62 & 0.1667 & 0.0833 & 0.1667 & - & 43.99 \\
    B\_Mime & 0.38 & 92.95 & 0.0000 & 0.0723 & 0.2474 & - & 38.66 \\
    B\_MeanSeq & 0.01 & 98.39 & 0.0000 & 0.0000 & 0.0000 & - & 45.39 \\
    B\_MeanFr & 0.00 & 99.04 & 0.0000 & 0.0000 & 0.0000 & - & 49.00 \\
    \midrule
    Trans-VAE w/o visual modality & 0.08 & 99.03  & 0.0229  &0.0029 & 0.0255 &  65.18 & 44.47 \\

    Trans-VAE w/o audio modality &0.09 & 96.83 & 0.0088  & 0.0013 & 0.0094 & 63.77 &45.24\\
    \midrule
    Trans-VAE   & 0.10  & 98.48  &  0.0242 & 0.0040 & 0.0263 & 69.24
    & 44.88 \\
    BeLFusion ($k$=1) & 0.12 & 90.21 & 0.0085 & 0.0056 & 0.0103 & - & 44.95 \\
    BeLFusion ($k$=10) & 0.13 & 89.84 & 0.0137 & 0.0078 & 0.0149 & - & 45.02 \\
    BeLFusion ($k$=10) + Binarized AUs & 0.12 & 92.58 & 0.0322 & 0.0170 & 0.0337 & - & 49.00 \\

\bottomrule
\end{tabular}
\end{table*}

\begin{figure*}[t!]
    \centering
    \includegraphics[width=1.5\columnwidth]{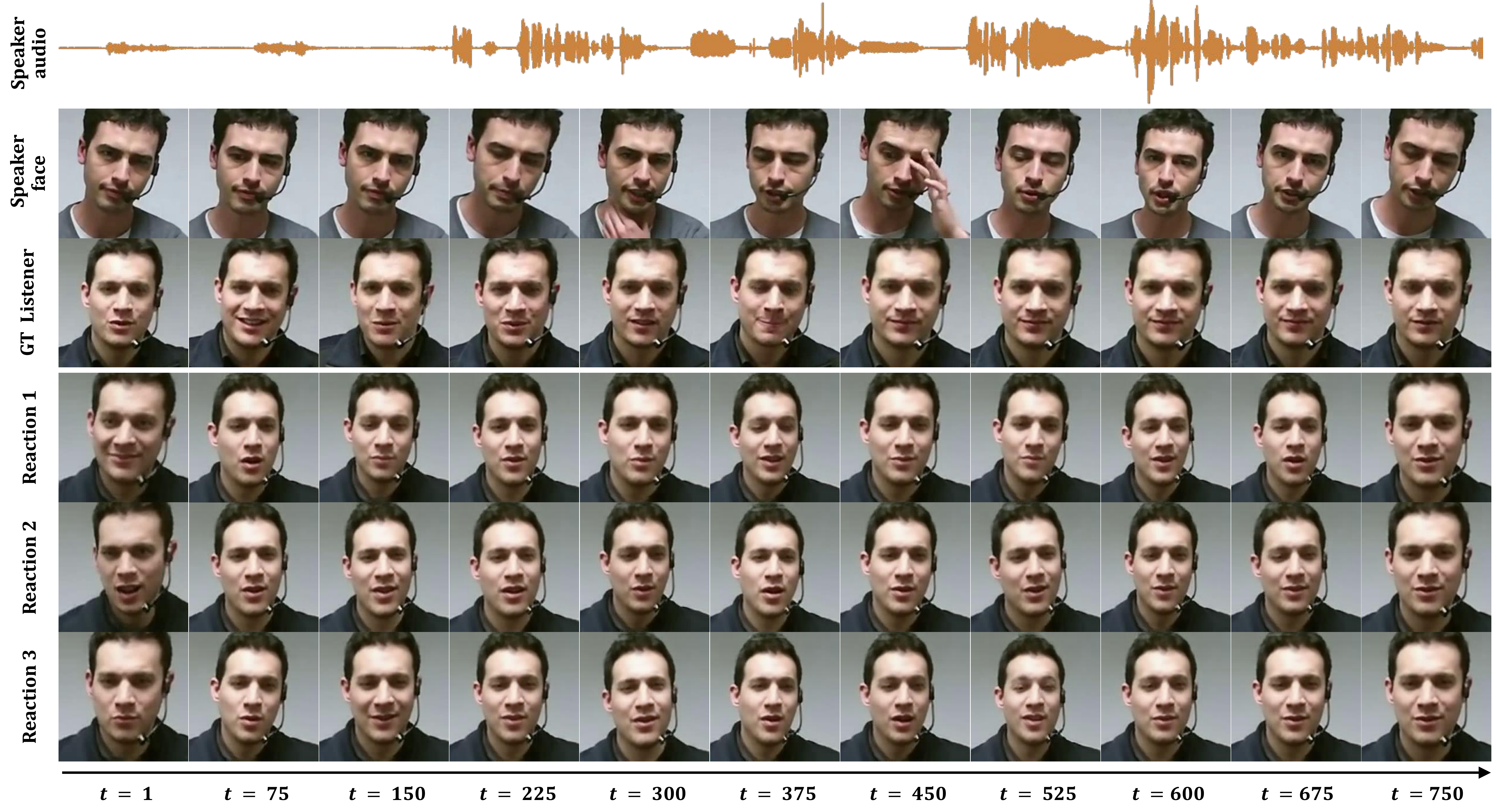} 
        \caption{Examples of generated multiple listener reactions to a given speaker behaviour (including the speaker's audio and face frames). These reactions are generated by an online Trans-VAE model.  }
    \label{fig:ex_vis_online}
\end{figure*}


\subsubsection{Offline facial reaction generation sub-challenge}

\autoref{tb:val_quantitative_offline} and \autoref{tb:test_quantitative_offline} show that both baselines can generate facial reactions that positively correlate to the appropriate real facial reactions, where BeLFusion outperforms Trans-VAE in terms of the distance between the prediction and real appropriate facial reactions (FRD), as well as intra-sequence diversity (FRVar). In return, the intra- and inter-subject diversities (S-MSE and FRDvs) for the non-binarized approaches are lower than Trans-VAE. Moreover, the results achieved by Trans-VAE suggest that both visual and audio modalities positively contribute to the diversity of generated facial reactions (FRDiv, FRVar, and FRDvs). As for the FRC metric, Trans-VAE shows the same limitations observed for the BeLFusion. For the latter, we observe that, as expected, higher values of $k$ boost all diversity metrics~\cite{barquero2022belfusion}. Randomly sampled facial reaction (i.e., B\_Random) are diverse but not appropriate (in terms of FRC and FRD), whereas deterministic baselines (i.e., B\_Mime, B\_MeanSeq and B\_MeanFr) achieved better appropriateness but much lower diversity. Unlike above baselines, deep learned probabilistic models (i.e., Trans-VAE and BeLFusion) can make a trade-off between the these two, which means they can generate multiple different but appropriate facial reactions (i.e., qualitative results achieved by the offline Trans-VAE are visualised in \autoref{fig:ex_vis_offline}).

\subsubsection{Online facial reaction generation sub-challenge}

As demonstrated in \autoref{tb:val_quantitative_online} and \autoref{tb:test_quantitative_online}, the results confirmed that both baselines can generate real-time facial reactions that are positively correlated to the appropriate face reactions. Also, Trans-VAE outperforms the BeLFusion approach in terms of diversity (FRDiv, FRVar, and FRDvs), synchrony (FRSync), and while BeLFusion outperforms in terms of DTW distances (FRD).  
Similarly to the offline task, the randomly sampled facial reactions (i.e., B\_Random) are diverse but not appropriate, while the deterministic naive approaches (i.e., B\_Mime, B\_MeanSeq and B\_MeanFR) achieved better results in terms of appropriateness but not diversity. Again, the proposed deep learning baselines can be better in trading-off between appropriateness and diversity.
In this sub-challenge, the differences are magnified for the four metrics. This again suggests the existence of a trade-off between the appropriateness and diversity of the generated facial reactions. In fact, such trade-off is fairly observed after binarizing the predicted AUs: while the FRD worsens, all diversity metrics are doubled. Compared to offline setting, online generation is more challenging and cause jitters and inconsistency between windows, which are main reasons for the decrease in  Realism (i.e., FRRea metric). However, online Trans-VAE approach outperformed the offline one. As a transformer-architecture network, it models a long-range relation between current and past reaction frames and attends to salient changes in speaker behaviours so as to achieve good synchrony in online scenario.
\autoref{fig:ex_vis_online} visualises that Trans-VAE can give real-time facial expression feedback to speaker behaviour in the online scenario and given reactions can be also multiple and diverse. 


\begin{table*}[t]\centering
  \footnotesize
  \caption{\label{tb:val_quantitative_online} 
 Baseline online facial reaction generation results achieved on the validation set.} 
\begin{tabular}{lccccccc}
\toprule 
\multirow{2}{*}{\textbf{Method}} & \multicolumn{2}{c}{\textbf{Appropriateness}}  & \multicolumn{3}{c}{\textbf{Diversity}} &\textbf{Realism}  &   \textbf{Synchrony}\\
\cmidrule(r){2-3} \cmidrule(r){4-6}  \cmidrule(r){7-7}  \cmidrule(r){8-8} 
    & \textbf{FRC} ($\uparrow$) &  \textbf{FRD} ($\downarrow$) &  \textbf{FRDiv} ($\uparrow$) &  \textbf{FRVar} ($\uparrow$) &  \textbf{FRDvs} ($\uparrow$)  & \textbf{FRRea} ($\downarrow$)    & \textbf{FRSyn} ($\downarrow$)\\ 
    \midrule
    GT & 8.42 & 0.00 & 0.0000 & 0.0666 & 0.2251 & 44.31 & 48.52 \\
    \midrule
    B\_Random & 0.04 & 229.37 & 0.1667 & 0.0833 & 0.1667 & - & 46.65 \\
    B\_Mime & 0.35 & 91.37 & 0.0000 & 0.0666 & 0.2251 & - & 44.53 \\
    B\_MeanSeq & 0.01 & 102.17 & 0.0000 & 0.0000 & 0.0000 & - & 47.19 \\
    B\_MeanFr & 0.00 & 102.84 & 0.0000 & 0.0000 & 0.0000 & - & 49.00 \\
    \midrule
    Trans-VAE w/o visual modality &  0.15 & 134.54 & 0.1141  & 0.0592 & 0.1220 & 90.39 & 49.00\\
    Trans-VAE w/o audio modality & 0.13 & 134.69 & 0.1090  & 0.0565 & 0.1134 &  97.40 & 49.00 \\
    \midrule
    Trans-VAE   & 0.14 & 134.40 &  0.1149 & 0.0594 & 0.1224 & 108.03
    &46.81 \\
    BeLFusion ($k$=1) & 0.12 & 93.91 & 0.0086 & 0.0058 & 0.0104 & - & 47.13 \\
    BeLFusion ($k$=10) & 0.14 & 93.64 & 0.0131 & 0.0076 & 0.0143 & - & 47.10 \\
    BeLFusion ($k$=10) + Binarized AUs & 0.12 & 96.55 & 0.0310 & 0.0167 & 0.0324 & - & 49.00 \\
\bottomrule
\end{tabular}
\end{table*}

\begin{table*}[t]\centering
 \footnotesize
 \caption{\label{tb:test_quantitative_online} 
 Baseline online facial reaction generation results achieved on the test set.} 
\begin{tabular}{lccccccc}
\toprule 
\multirow{2}{*}{\textbf{Method}} & \multicolumn{2}{c}{\textbf{Appropriateness}}  & \multicolumn{3}{c}{\textbf{Diversity}} &\textbf{Realism}  &   \textbf{Synchrony}\\
\cmidrule(r){2-3} \cmidrule(r){4-6}  \cmidrule(r){7-7}  \cmidrule(r){8-8} 
    & \textbf{FRC} ($\uparrow$) &  \textbf{FRD} ($\downarrow$) &  \textbf{FRDiv} ($\uparrow$) &  \textbf{FRVar} ($\uparrow$) &  \textbf{FRDvs} ($\uparrow$)  & \textbf{FRRea} ($\downarrow$)    & \textbf{FRSyn} ($\downarrow$)\\ 
    \midrule
    GT & 8.74 & 0.00 & 0.0000 & 0.0723 & 0.2474 & 47.50 & 47.72 \\
    \midrule
    B\_Random & 0.04 & 237.62 & 0.1667 & 0.0833 & 0.1667 & - & 43.99 \\
    B\_Mime & 0.38 & 92.95 & 0.0000 & 0.0723 & 0.2474 & - & 38.66 \\
    B\_MeanSeq & 0.01 & 98.39 & 0.0000 & 0.0000 & 0.0000 & - & 45.39 \\
    B\_MeanFr & 0.00 & 99.04 & 0.0000 & 0.0000 & 0.0000 & - & 49.00 \\
    \midrule
    Trans-VAE w/o visual modality & 0.13 & 134.78 & 0.1121  & 0.0581 & 0.1166 & 69.20 & 44.24 \\
    Trans-VAE w/o audio modality &  0.13 & 134.77  & 0.1087  & 0.0564 & 0.1095 & 74.54 & 44.33 \\ \midrule
    
    Trans-VAE   & 0.13 & 135.57 & 0.1168 & 0.0604 & 0.1202 & 71.15 & 44.31\\
    BeLFusion ($k$=1) & 0.12 & 89.56 & 0.0086 & 0.0058 & 0.0103 & - & 45.09 \\ 
    BeLFusion ($k$=10) & 0.13 & 89.42 & 0.0133 & 0.0077 & 0.0143 & - & 44.80 \\ 
    BeLFusion ($k$=10) + Binarized AUs & 0.12 & 92.13 & 0.0306 & 0.0164 & 0.0317 & - & 49.00 \\
\bottomrule
\end{tabular}
\end{table*}




\section{Conclusion}
\label{sec: conclusion}

In this paper, we introduced REACT2023 - the first Multiple Appropriate Facial Reaction Generation challenge, which provides the very first attempt to bring together researchers from different subjects to contribute a new challenging but promising affective computing research direction. It comprises two sub-challenges: (i) Offline Multiple Appropriate Facial Reaction Generation challenge; and (ii) Online Multiple Appropriate Facial Reaction Generation challenge. Intentionally, we provide not only audio-visual dyadic interaction clips that segmented from three different datasets with various interaction conditions, but also a set of audio-visual baseline features extracted from open-source software/code with the highest possible transparency and realism for the baselines. Importantly, we made all the code scripts for both features extraction and two facial reaction generation baselines (i.e., Trans-VAE and BeLFusion) to be publicly available, where both baselines can generate multiple different but appropriate and realistic facial reactions from speaker audio-visual behaviours in both offline and online settings. Our protocol strictly evaluates all participant models under the same settings by comprehensively considering four aspects of their generated facial reactions: appropriateness, diversity, realism and synchrony. 

The results of our proposed baselines suggested that:
(i) both Trans-VAE and BeLFusion baselines achieved better results in making a trade-off between appropriateness and diversity of the facial reactions with respect to the naive baselines;
(ii) both visual and audio modalities in Trans-VAE positively contributed to the diversity and appropriateness of the generated facial reactions;
(iii) the Trans-VAE achieved better results for online sub-challenge over the offline sub-challenge, while the opposite scenario has been observed in the BeLFusion approach;


As the first multiple appropriate facial reaction generation challenge, the used dataset is not specifically recorded, and thus some important behavioural cues (e.g., verbal texts, physiological signals) were not considered in this challenge. Our future work will focus on continue organizing REACT challenges while introducing a new dataset containing more modalities.

\section*{Acknowledgements}
\textbf{Funding:} M. Spitale and H. Gunes are supported by the EPSRC/UKRI under grant ref. EP/R030782/1 (ARoEQ). This work has been partially supported by the Spanish project PID2019-105093GB-I00 and by ICREA under the ICREA Academia programme.

\noindent \textbf{Open Access:} For open access purposes, the authors have applied a Creative Commons Attribution (CC BY) licence to any Author Accepted Manuscript version arising.

\noindent \textbf{Data access:} Data related to this publication can be accessed upon request following the terms and conditions of the datasets' owners.

\clearpage

\bibliographystyle{ACM-Reference-Format}
\balance
\bibliography{sample-base}

\end{document}